\documentclass[conference]{IEEEtran}
\IEEEoverridecommandlockouts
\usepackage{cite}
\usepackage{amsmath,amssymb,amsfonts}
\usepackage{algorithmic}
\usepackage{graphicx}
\usepackage{textcomp}
\usepackage{xcolor}
\usepackage{framed,multirow}
\usepackage{multirow}
\usepackage{adjustbox}
\usepackage{floatrow}
\usepackage{booktabs}
\usepackage{amssymb}
\usepackage{latexsym}
\usepackage{amsmath}
\usepackage{graphicx}
\usepackage{adjustbox}
\usepackage{url}
\usepackage{xcolor}
\usepackage{cellspace}
\usepackage{makecell}
\usepackage{comment}
\usepackage{amssymb}
\usepackage{latexsym}
\usepackage{adjustbox}
\usepackage{makecell}
\usepackage[caption=false, font=footnotesize]{subfig}

\usepackage{url}

\def\BibTeX{{\rm B\kern-.05em{\sc i\kern-.025em b}\kern-.08em
    T\kern-.1667em\lower.7ex\hbox{E}\kern-.125emX}}
\begin{document}

\title{Pose-GNN : Camera Pose Estimation System Using Graph Neural Networks
\thanks{This paper is under review by Computer vision and image understanding journal}
}
%\author{\uppercase{Ahmed M. Elmoogy}\authorrefmark{1},
%\uppercase{Xiaodai Dong\authorrefmark{1}}, \uppercase{TAO LU\authorrefmark{1}}, \uppercase{Robert Westendorp\authorrefmark{2} and} \uppercase{Kishore Reddy\authorrefmark{2}}}

\author{\IEEEauthorblockN{Ahmed Elmoogy\IEEEauthorrefmark{1}, Xiaodai Dong\IEEEauthorrefmark{1}, Tao Lu\IEEEauthorrefmark{1}, Robert Westendorp\IEEEauthorrefmark{2} and Kishore Reddy\IEEEauthorrefmark{2}} 
\IEEEauthorblockA{\IEEEauthorrefmark{1}Electrical and computer engineering, University of Victoria,Victoria BC, Canada (e-mail: \{ahmedelmoogy,xdong,taolu\}@uvic.ca)} \IEEEauthorblockA{\IEEEauthorrefmark{2}Fortinet, Burnaby BC, Canada (e-mail: \{rwestendorp,kreddy\}@fortinet.com)}}

\maketitle

\begin{abstract}
We propose a novel image based localization system using graph neural networks (GNN). The pretrained ResNet50 convolutional neural network (CNN) architecture is used to extract the important features for each image. Following, the extracted features are input to GNN to find the pose of each image by either using the image features as a node in a graph and formulate the pose estimation problem as node pose regression or modelling the image features themselves as a graph and the problem becomes graph pose regression. We do an extensive comparison between the proposed two approaches and the state of the art single image localization methods and show that using GNN leads to enhanced performance for both indoor and outdoor environments. 
\end{abstract}

\section{Introduction}
Camera pose estimation is one of the common methods for localizing an agent in any environment. It is a part of the simultaneous localization and mapping (SLAM) system where an agent can be equipped with multiple sensors and use them to localize itself in an unknown environment and build a map simultaneously. Camera pose estimation can be also used for relocalization if the agent loses tracking of its pose and needs to localize itself again which requires the system to be accurate and simple. Classical camera pose estimation systems rely on either 2D-2D or 2D-3D matchings. Hand crafted features such as speeded-up robust features (SURF)  \cite{bay2008speeded} and  oriented fast rotated brief (ORB) \cite{rublee2011orb} of the consecutive images are matched to find the relative pose between them \cite{mur2015orb}. For 2D-3D methods, a 3D model of the environment built using structure from motion (sfm) \cite{ullman1979interpretation} or point clouds is used to match between the 2D image features and the 3D model and the pose is estimated using n-point algorithm \cite{hartley1997defense} inside a RANSAC loop \cite{choi1997performance}.
To enhance the performance of the pose estimation, machine learning techniques are used without the need for extracting hand crafted features or matching the images which is susceptible to errors and drift. Neural networks are used for the pose estimation tasks by either directly regressing the pose of the single image or by regressing scene coordinated by constructing the depth map of every pixel. The later methods can lead to similar or better results compared to the classical pose estimation methods but they require very complex training procedure and a ground truth depth map during training. 

Our work belongs to the single image pose estimation system, however, for the first time according to the authors’ knowledge, to use the graph neural network (GNN) for image pose estimation. GNN is a class of neural networks that learn to do tasks by aggregating the features of graph structured data. A graph $G = (V,E)$ is defined as a set of nodes or vertices $V$ and connections between these nodes or edges $E$. Each node is associated with features which is processed initially by GNN. The purpose of GNN is to update the node represention through multiple layers by aggregating the information of the neighbors of every node through message passing to have final hidden features for every node \cite{wu2020comprehensive}. Nodes representation can be updated using recurrent neural networks (RecGNN) \cite{li2015gated} or using multiple forms of convolution (ConvGNN) \cite{duvenaud2015convolutional}. The final node representation can be used for many tasks including node level tasks where there is a label for every node such as semi supervised node classification \cite{kipf2016semi} or graph level tasks where there is one label for the whole graph such as molecule property regression \cite{gilmer2017neural}.

\subsection{Contribution}
Single image pose estimation networks mostly use pretrained convolutional neural networks %with pretrained initialization 
with the addition of more layers and use transfer learning to train the overall structure in an end to end fashion. We propose a new pipeline to this process. Firstly, instead of training the pretrained CNN again, we show that the pretrained features are good enough for pose estimation task without further fine-tuning. Secondly, instead of using fully connected layers \cite{kendall2015posenet} or LSTM layers \cite{walch2017image}, we leverage the power of GNN to process the pretrained features. We propose two novel approaches to model the image features with ConvGNN: 
\begin{enumerate}
    \item Image as a node (node-pose).
    \item Image as a graph (graph-pose).
\end{enumerate}
For the node-pose approach, the second to last layer features vector of pretrained ResNet50  \cite{he2016deep} are extracted for every image which will be used as the representation of every image as a node in a big graph which contains connections between neighboring images based on the similarity between the pretrained features. The ConvGNN is trained to regress the pose of every node in the graph.
The second approach deals with image as a graph where instead of extracting the flattend pretrained features, the intermediate feature maps of ResNet50 are extracted for each image and then converted into a graph structured data by connecting the nearest neighbors of the features and ConvGNN is used to to regress the pose of each graph.

We do extensive experiments using both indoor and outdoor datasets and show the superiority of both systems compared to the state of the art and the advantages of using either approach for pose estimation. 
\section{Related Work}
In feature based image localization methods, features are extracted and matched using feature detectors and descriptors.  ORB descriptors are used in \cite{mur2015orb,mur2017orb}  to find the  correspondences between consecutive frames and then using 8-point algorithm to find the relative poses. Other techniques  \cite{sattler2011fast, sattler2017efficient} use the 2D-3D correspondence between the 2D image and the 3D model using, e.g., structure from motion (sfm) to solve finding pose problem \cite{forsyth2011computer}. 
For direct methods, the whole image information is used in  \cite{engel2014lsd} which leads to a denser map. In addition, the work \cite{sattler2015hyperpoints} uses image retrieval techniques to search for the similarity between  the current image and images in the database, then find the relative poses. 
Learning-based SLAM techniques are being enhanced day after day, especially for pose estimation problems. For example, transfer learning ~\cite{oquab2014learning} and the pretrained GoogLeNet~\cite{szegedy2015going} are used in \cite{kendall2015posenet} to find the absolute pose of the camera. Similar architecture  were implemented in \cite{hazirbasimage} with long-short term memory (LSTM) \cite{schmidhuber1997long} to memorize good features which help finding the absolute pose. Further, pretrained  ResNet-34 ~\cite{he2016deep} is applied for regressing camera pose in \cite{melekhov2017image} by  adopting encoder-decoder architecture with skipped connections to move the features from the early layers to the output layers.   CNN architecture and image retreival techniques are combined in \cite{iyer2018geometric} in which the proposed model contains two subnetworks, the representation part which learns to represent the image with fixed length vector for the image retrieval task and the regression subnetwork which finds the relative pose between the query image and the retrieved image. VidLOC \cite{clark2017vidloc} proposes a video localization architecture using the combination of CNN and Bidirectional LSTM to find the absolute pose of each image in videos with different lengths. Finally, to reduce the training complexity, many ideas are proposed using SURF descriptors with both CNN and LSTM \cite{elmoogy2020surfcnn}, \cite{elmoogy2020surf_lstm}, \cite{elmoogy2020generalizable} and Linear regression \cite{elmoogy2020linear} to learn the pose of single images. 
\section{Graph Neural network for image pose estimation}
\subsection{Convolutional graph neural network (ConvGNN)}
ConvGNN relies on doing convolutions on the sparse graph structure which is different from the image convolution with defined grid structure. Convolution is usually done in two ways: 
\begin{enumerate}
    \item Spectral convolution (Spectral ConvGNN)
    \item Spatial convolution (Spatial ConvGNN)
\end{enumerate}
\begin{figure*}[t]
\centering
\includegraphics[scale=.4]{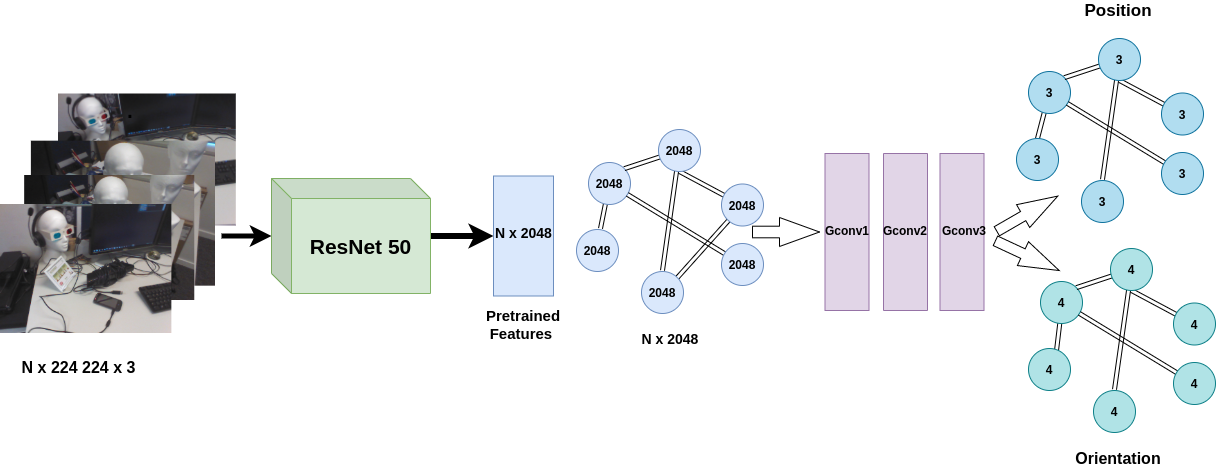}
\caption{Image as a node architecture.}
\label{node-pose}
\end{figure*}

\subsubsection{Spectral ConvGNN} 
For undirected graph with adjacency matrix $A$ and a degree matrix $D$ which is the sum of neighbors to every node $D_i = \sum_j A_{ij}$, we can use the normalized graph Laplacian matrix $L = I_n + D^{-0.5}AD^{-0.5} $ to approximate the graph convolution using Fourier transform properties \cite{wu2020comprehensive}. There are many implementations and approximations to spectral ConvGNN such as GCN \cite{kipf2016semi} and ChebNet \cite{monti2017geometric}. In our experiments, we use GCN for its simplicity and proven performance in multiple applications \cite{zhou2018graph}. GCN follows the node representation update as 
\begin{equation}
    X^{'} = \hat{D}^{-0.5}\hat{A}\hat{D}^{-0.5}X\theta
    \label{EQ:GCN}
\end{equation}
where $X^{'}$ is the new nodes features representation, X is the current nodes features, $\hat{A} = A + I_n$ is the adjacency matrix with added self loops (node is connected to itself), $\hat{D}$ is the node degree matrix where $\hat{D}_{i} = \sum_j \hat{A}_{ij}$ and $\theta$ is the learning parameters or the weights.  

\subsubsection{Spatial ConvGNN}
Instead of relying on the spectral theory to approximate the convolution operation, spatial methods try to use the same convention of convolution done on regular grid structures like images which is simple aggregation of neighbors such as addition, product, minimum and maximum. Many spatial ConvGNN implement different aggregation criteria including GraphSage \cite{hamilton2017inductive}, Graph attention networks (GAT) \cite{velivckovic2017graph} and Weisfeiler and Leman-GNN (WL-GNN) \cite{morris2019weisfeiler}. We use WL-GNN for implementing our systems as it has a different set of weights for the central node than the neighbors which helps improve the performance according to \cite{dwivedi2020benchmarking}. WL-GNN has the following node update formula 
\begin{equation}
    x^{'}_i = \theta_1 x_i + \sum_{j \in N(i)} \theta_2 x_j
\end{equation}
where $x^{'}_i$ and $x_i$ are the new and current representation of node $i$, $\theta_1$ and $\theta_2$ are the weights for the central node $i$ and the neighbors $j$ to node $i$ lying in the set $N(i)$ of all neighbors to node $i$.  
\label{GNN_types}
\subsection{Image as a node (node-pose)}
Fig. \ref{node-pose} shows the details of the first architecture where images are modeled as nodes in a graph. To designate an image as a node, each image should be represented by a feature vector which is used as the initial features of the node. CNN architectures that are trained on a big image classification dataset for days are proven to act as a backbone to many applications, providing an initial feature set to start with. Here, we use ResNet50 CNN architecture \cite{he2016deep} that is widely used in many applications. Images are input to pretrained ResNet50 and the output of the second to last layer is taken as the feature vector of size $2048$ for each image. 
\subsubsection{Training}
For $N$ training images with size $(N\times 224 \times 224 \times 3)$, pretrained features are extracted to a total size $(N\times2048)$. Next, the pretrained features representation is converted into a graph structure by building connections (edges) between images in the form of binary adjacency matrix. We use K-nearest neighbors (KNN) algorithm \cite{peterson2009k} to search for the nearest K neighbors of every image based on the $L_2$ distance between every image's pretrained features. For two images $I_i, I_j$ with pretrained feature representation $x_i$ and $x_j$, the $L_2$ distance is calculated as 
\begin{equation}
    d(I_i,I_j) = d(x_i, x_j) = \sqrt{\sum_n (x_{in} - x_{jn})^2}
    \label{L2}
\end{equation}
where $n$ is the dimension of the pretrained features, i.e., $2048$. The images with the smallest $K$ distances are assigned as neighbors. Following, we use the feature matrix of the images $X \in R^{N \times 2048}$ and the binary adjacency matrix %calculated using the KNN 
$A \in R^{N \times N}$ as input to the GNN. There are 3 graph convolutional layers (Gconv). The first layer accepts the node feature representation of size $2048$ and learn new features of size $256$ by aggregating the neighboring features of each node according to (\ref{EQ:GCN}). Following, new hierarchical features are learned in Gconv2 and Gconv3 of sizes $128$ and $64$ respectively. The output of the last Gconv layer (Gconv3) is fed to the final fully-connected layers of size $3$ and $4$ for position $p = [x, y, z]$ and orientation $q = [q_w, q_x, q_y, q_z]$ in quaternion form. The $L_2$ loss function to learn the pose of the images is given by 
\begin{equation}
loss(X) = ||p\hat{} - p||_2 + \alpha||q\hat{} - q||_2 
\label{loss}
\end{equation}
where $\alpha$ is used to balance the scaling between position and orientation. We choose the value of $\alpha$ to be $200$ for outdoor environments and $10$ for indoor environments. 
\subsubsection{Testing}
After training is done and the weights of the model are saved, testing can be done for single or multiple images different than the training data. Specifically, for one or multiple testing images, pretrained features are extracted with size $2048$. Then,  the K-nearest images from the training image set are identified to construct a new graph containing the testing images and the K nearest training images by finding the $L_2$ distance between the pretrained features expressed in (\ref{L2}) . Finally, the saved model is used to aggregate information from the testing images and their neighbors and regresses the poses of the testing images. 

\begin{figure*}[t]
\centering
\includegraphics[scale=.4]{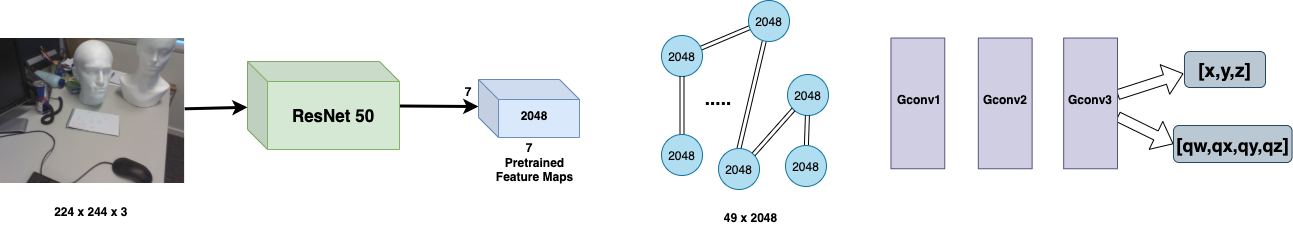}
\caption{Image as a graph architecture.}
\label{image_as_a graph}
\end{figure*}

\subsection{Image as a graph (graph-pose)}
Here, instead of dealing with images as nodes in a graph, we represent the images themselves as graphs. So if we have $N$ images, we will have $N$ small graphs instead of one big graph with $N$ nodes. To represent the image as a graph, we need to decompose the image into a set of nodes and construct connections or edges between them. Dealing with the original image is not a feasible solution as it contains thousands of pixels that can be candidate nodes. In our system, we use ResNet50 pretrained CNN to extract feature nodes of the input image. All the images are input to ResNet50, then, instead of extracting the second to last flattened features, we extract the intermediate feature maps with general size $L \times W \times d$ denoting for the length, width and depth. These feature maps result from downsampling the input image through convolution, pooling and nonlinear activation function at the different layers with different number of convolutional filters. The details of the graph-pose system are shown in Fig. \ref{image_as_a graph} where the input image is fed to ResNet50 and feature maps of the intermediate layers are extracted. ResNet50 consists of 50 layers: $1$ flattened layer and $49$ different feature maps that can be used as input to GNN. We use the feature maps from the layer before the flattened features with size $7 \times 7 \times 2048$. These feature maps can be reshaped to $49 \times 2048$ and be represented as a graph that contains $49$ nodes with features vector of size $2048$ for each of them. Next, edges between different nodes are constructed using the K-nearest neighbors with $L_2$ distance criteria between different nodes. For training this system, we use the same $L_2$ loss in (\ref{loss}).

\begin{figure*}[!t]
\centering
\includegraphics[scale=.35]{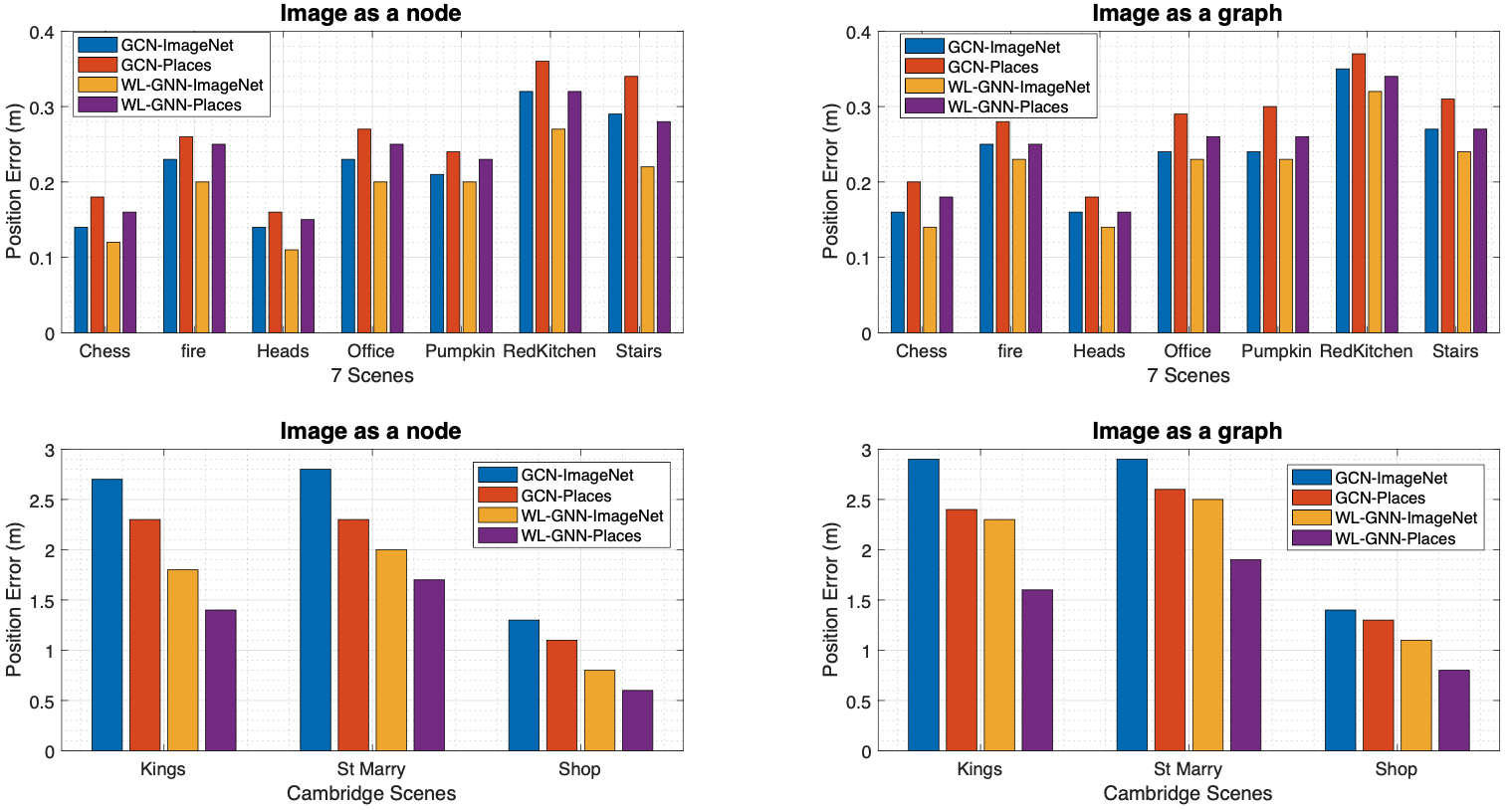}
\caption{The effect of pretrained features and type of GNN on the position error for the 7 scenes and Cambridge datasets.}
\label{pretrained}
\end{figure*}

\begin{figure*}
    \centering
  \subfloat[Heads Scene\label{1a_h}]{%
       \includegraphics[width=0.49\linewidth]{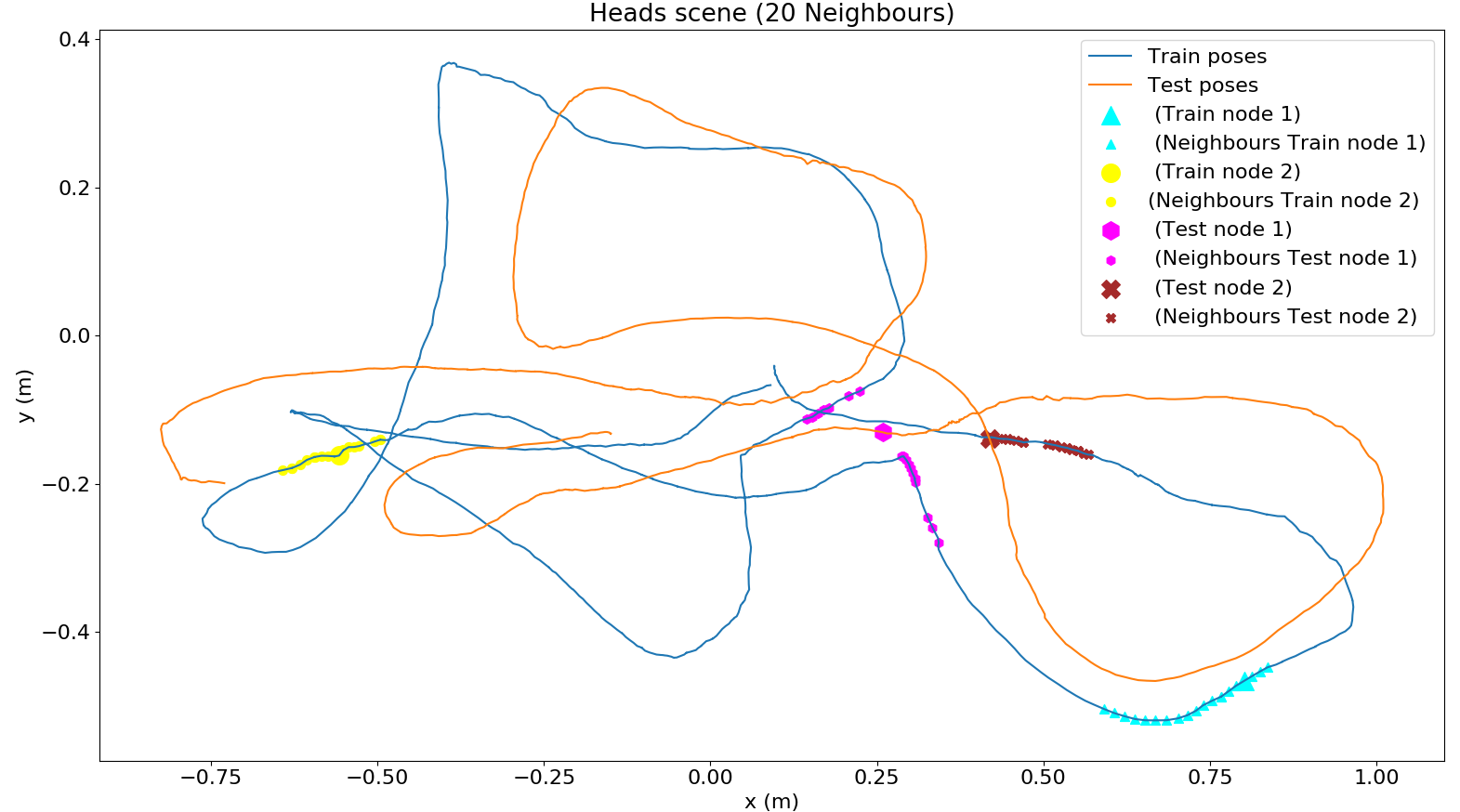}}
   \quad
  \subfloat[Kings Scene\label{1b_k}]{%
        \includegraphics[width=0.49\linewidth]{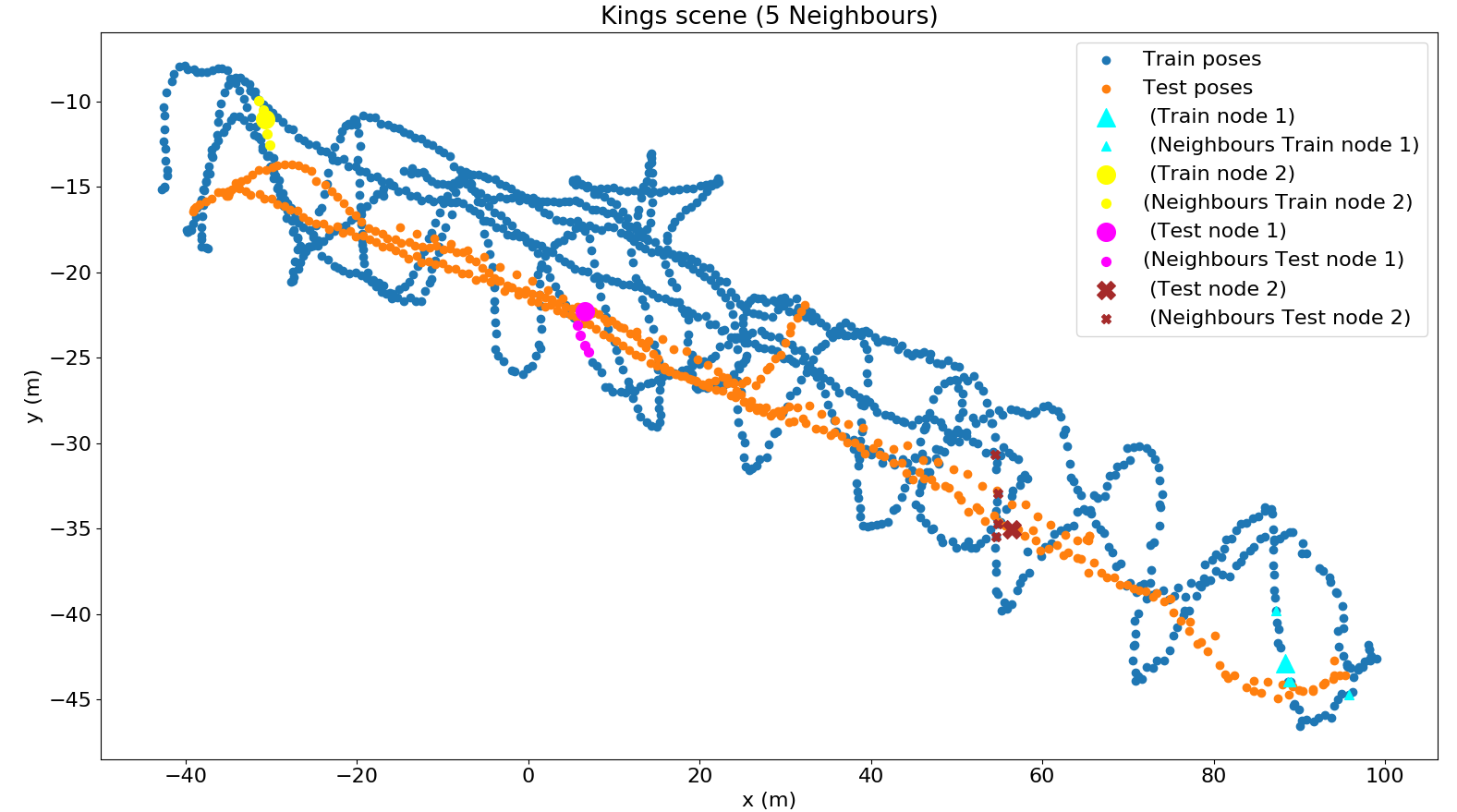}}

  \caption{The connected positions for multiple training and testing nodes of Heads scene (left) and Kings scene (right).}
  \label{neigbours_node} 
\end{figure*}

\section{Datasets}
For testing the proposed systems in different conditions, we use one indoor dataset and one outdoor dataset:
\begin{enumerate}
    \item The 7 scenes dataset \cite{glocker2013real} consists of 7 different scenes ranging from small area, low field of views images to wide areas and complex scenes. Each scene contains the training images and testing images with their ground truth poses. 
    \item The Cambridge dataset \cite{kendall2017geometric} contains multiple scenes from outdoor environments with multiple images for training and testing alongwith the ground truth poses. 
\end{enumerate}
\section{Design Analysis}
In this section, we discuss the effect of multiple design consideration on the localization accuracy including the choices of the pretrained dataset, the type of ConvGNN and the number of neighbors.
\subsection{Pretrained features and types of ConvGNN}
For both node-pose and graph-pose architectures, we employ ResNet50 CNN to extract the features of the input images which is pretrained on large image dataset. ImageNet dataset \cite{russakovsky2015imagenet}  and Places dataset \cite{zhou2016places} are used as two pretrained datasets and study the effect of using them on the positional error of both the 7 scenes and Cambridge datasets. The position error of two types of ConvGNN, spectral ConvGNN (GCN) and spatial ConvGNN (WL-GNN), is shown in Fig. \ref{pretrained}. Multiple observations can be concluded from Fig. \ref{pretrained}: 
\begin{itemize}
    \item Using ImageNet dataset always produces better results for indoor scenes than Places dataset. 
    \item Using Places dataset gives lower error for outdoor scenes. 
    \item WL-GNN outperforms GCN across all scenes and for both systems. 
    \item Node-pose architecture performs better than graph-pose system. 
\end{itemize}
\label{observations}
Here, we discuss the above-mentioned observations in details. Firstly, ImageNet dataset is an object classification dataset that contains mostly one type of objects with simple scenes while Places dataset is a scene classification dataset with more complex scenes, many objects and mostly outdoor images. This can be one reason why ImageNet dataset pretrained features perform well for indoor scenes where the images contain few objects and scenes are not complex. On the other hand, Places dataset pretrained features work well for outdoor scenes where the scenes are more complex and the images contain many objects.  Secondly, we observe the superiority of WL-GNN over GCN for all scenes. One big difference between WL-GNN and GCN is that WL-GNN gives a different set of weights to the central node than the neighbors in contrast to GCN that applies the same set of weights to all nodes. As mentioned in the benchmark paper \cite{dwivedi2020benchmarking}, learning separate weights for central nodes is beneficial to the learning process and provide better results for multiple tasks and datasets. Finally, node-pose system provides better results since multiple images are used as neighbors for the final decisions while only the current image information is used by graph-pose architecture without connecting to neighboring images.

\begin{table*}[t]
\caption{The median error in position (m)/ orientation (degrees) for the 7 scenes and Cambridge dataset for the proposed models, compared with SurfCNN \cite{elmoogy2020surfcnn}, SURF-LSTM \cite{SURFLSTM}, PoseNet~\cite{kendall2015posenet}, G-Posenet~\cite{cai2018hybrid}, Posenet-U~\cite{kendall2016modelling}, Pose-L~\cite{hazirbasimage}, \cite{wu2017delving} and  G-PoseNet ~\cite{kendall2017geometric}, BranchNet \cite{wu2017delving} ,Mobile-PoseNet \cite{cimarelli2019faster},VidLoc \cite{clark2017vidloc} and Pose-Hourglass \cite{melekhov2017image}}
\large
\label{Median Error}
\centering
\begin{adjustbox}{width=1\textwidth}

\begin{tabular}{|l|c|c|c|c|c|c|c|c|c|c|c|c|} 
\hline
Algorithm& Chess & Fire & Heads & Office & Pumpkin & Kitchen & Stairs & Average & Kings & St Marry & Shop  & Average  \\ 
\hline
 \makecell{Node-Pose \\(ImageNet)}& 0.12/6&0.20/8.4&0.11/11.1&0.20/6.8&0.20/5.5&0.27/7.5&0.22/7.2&0.18/7.5&1.8/3.5&2/5.5&0.8/47&1.46/4.16 \\
\hline
\makecell{Node-Pose \\ (Places) } & 0.16/8 &0.25/9.4&0.15/11.8&0.25/8.1&0.23/6.5&0.32/9.5&0.28/9.2&0.23/8.9&1.4/2.8&1.7/5&0.6/3.57&1.2/3.76\\

\hline

\makecell{Graph-Pose\\ (ImageNet)} &0.14/6.4&0.23/8.6&0.15/11.8&0.23/6.8&0.23/5.5&0.32/7.7&0.24/7.4&0.22/7.84&2.3/4.6&2.5/6.9&1.1/5.1&1.9/5.53\\
\hline

\makecell{Graph-Pose \\(Places)} & 0.18/8.1&0.25/8.8&0.15/11.9&0.26/8.9&0.26/5.5&0.34/10&0.27/9.47&0.25/9&1.6/3.1&1.9/5.5&0.8/4&1.43/4.2
  \\

\hline

SURF-CNN & 0.19/8.10 & 0.24/12.8 & 0.17/12 & 0.35/9 & 0.36/10.8 &0.39/10.2 &0.36/10.8 &0.3/10.22 &7.66/12.3 &4.5/10.5 &3.2/9.2 &6.82/10.6  \\

\hline 

SURF-LSTM & 0.22/7.04 &0.24/10.2 &0.16/13.6  &0.35/8.5 &0.35/7.8 &0.38/9.5 &0.35/11.9 & 0.3/9.4& 5.53/10.6 & 3.2/9.2 & 1.46/8.1 & 5.37/9.3  \\

\hline

PoseNet & 0.32/8.12 &0.47/14.4 &0.29/12.0 & 0.48/8.42 &0.47/8.42 &0.59/8.64 &0.47/13.8 &0.44/11.63 &1.92/5.40 &1.46/8.1 &1.11/7.6 & 2.01/7.03 \\

\hline

Dense VLAD&  0.21/12.5& 0.33/13.8& 0.15/14.9& 0.28/11.2& 0.31/11.3 &0.30/12.3& 0.25/15.8 &0.26/13.11& 2.80/5.75 &1.11/7.6 &1.25/7.5 &2.07/6.95 \\

\hline
Bay. PN & 0.37/7.24 &0.43/13.7 &0.31/12.0 &0.48/8.04&0.61/7.08 &0.58/7.54&0.48/13.1& 0.46/9.81 &1.74/4.06 &1.25/7.5 & 1.79/6.5 &  1.70/6.2 \\ 

\hline

PN L-W & 0.14/4.50 & 0.27/11.8 & 0.18/12.1 & 0.20/5.77 & 0.25/4.84 & 0.24/5.52 & 0.37/10.6 & 0.24/7.87& 0.99/1.06 & 1.05/3.9 & 1.18/7.4 & 1.17/4.12  \\ 

\hline

LSTM. PN & 0.24/5.77  &0.34/11.9 & 0.21/13.7 & 0.30/8.08 & 0.33/7.00  &0.37/8.83  &0.40/13.7 & 0.31/9.85 &0.88/1.04 &1.18/7.4 & 1.14/5.7 &  1.19/4.71  \\

\hline
G PoseNet & 0.20/7.11 & 0.38/12.3 & 0.21/13.8 & 0.28/8.83 & 0.37/6.94 & 0.35/8.15 &0.37/12.5 & 0.31/9.94 & 0.99/3.65 & 1.14/5.7 &3.2/9.2 & 1.68/6.18  \\

\hline
Pose-Hourglass & 0.15/6.17 &0.27/10.8 &0.19/11.6& 0.21/8.48 &0.25/7.01 &0.27/10.2 &0.29/12.5& 0.23/9.5&NA&NA&NA&NA\\
\hline

Mobile-PoseNet  & 0.19/8.22 & 0.37/13.2 & 0.18/15.5 & 0.27/8.54 & 0.34 /8.46 &0.31/8.05 & 0.45/13.6 & 0.30/10.79 & 1.14/1.53 & 2.18/6.1 & 1.73/6.19 & 1.6/4.60  \\ 

\hline 

Vidloc &0.16 &0.21 & 0.14  &0.24 &0.36 & 0.31 & 0.26 & 0.25 & NA & NA & NA & NA \\ 
\hline
\end{tabular}
\end{adjustbox}

    \label{table: 7_scenes}
\end{table*}

\begin{figure*}[h]
    \centering
  \subfloat[Image as a node\label{1a}]{%
       \includegraphics[width=0.49\linewidth]{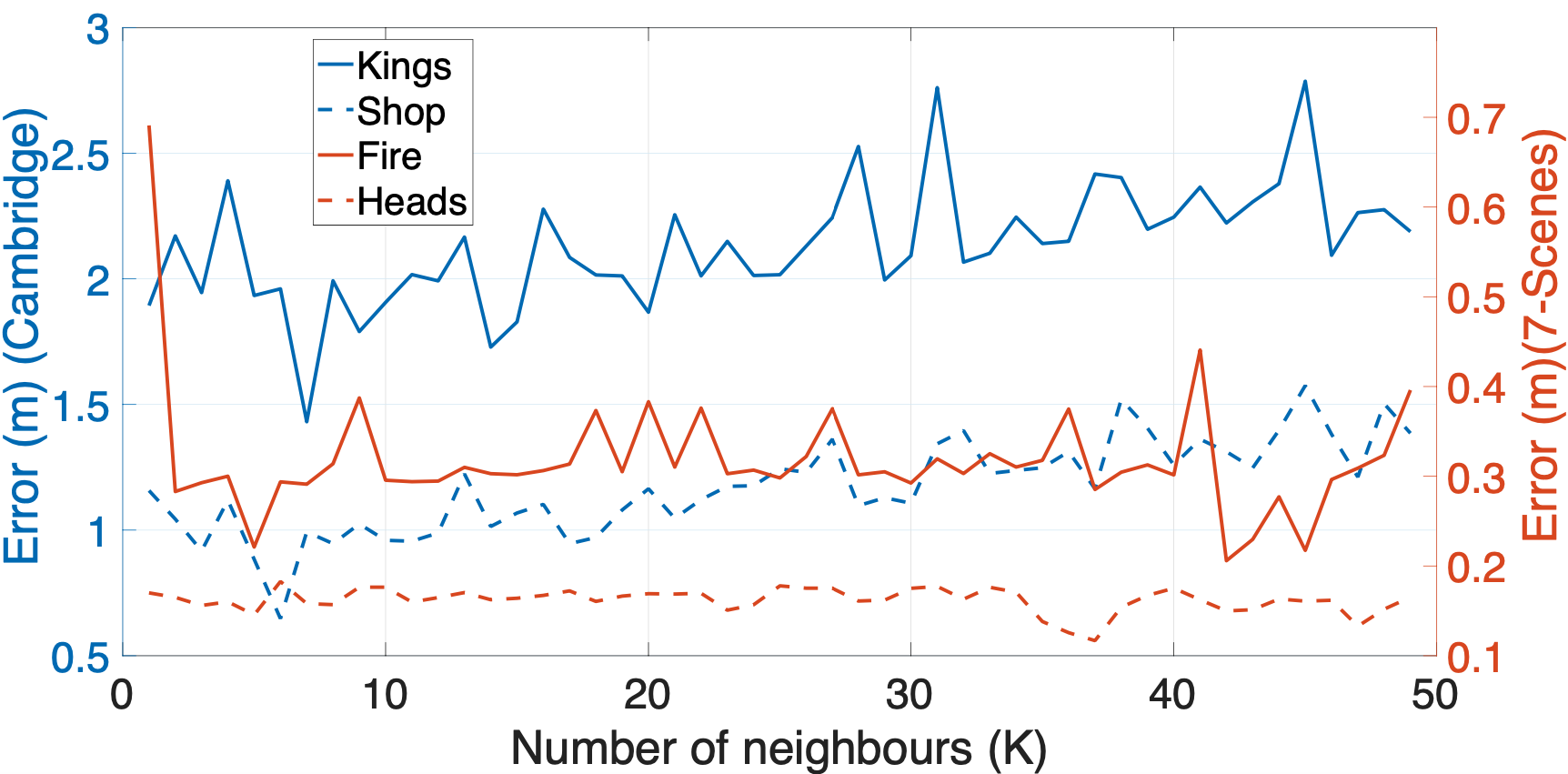}}
   \quad
  \subfloat[Image as a graph\label{1b}]{%
        \includegraphics[width=0.49\linewidth]{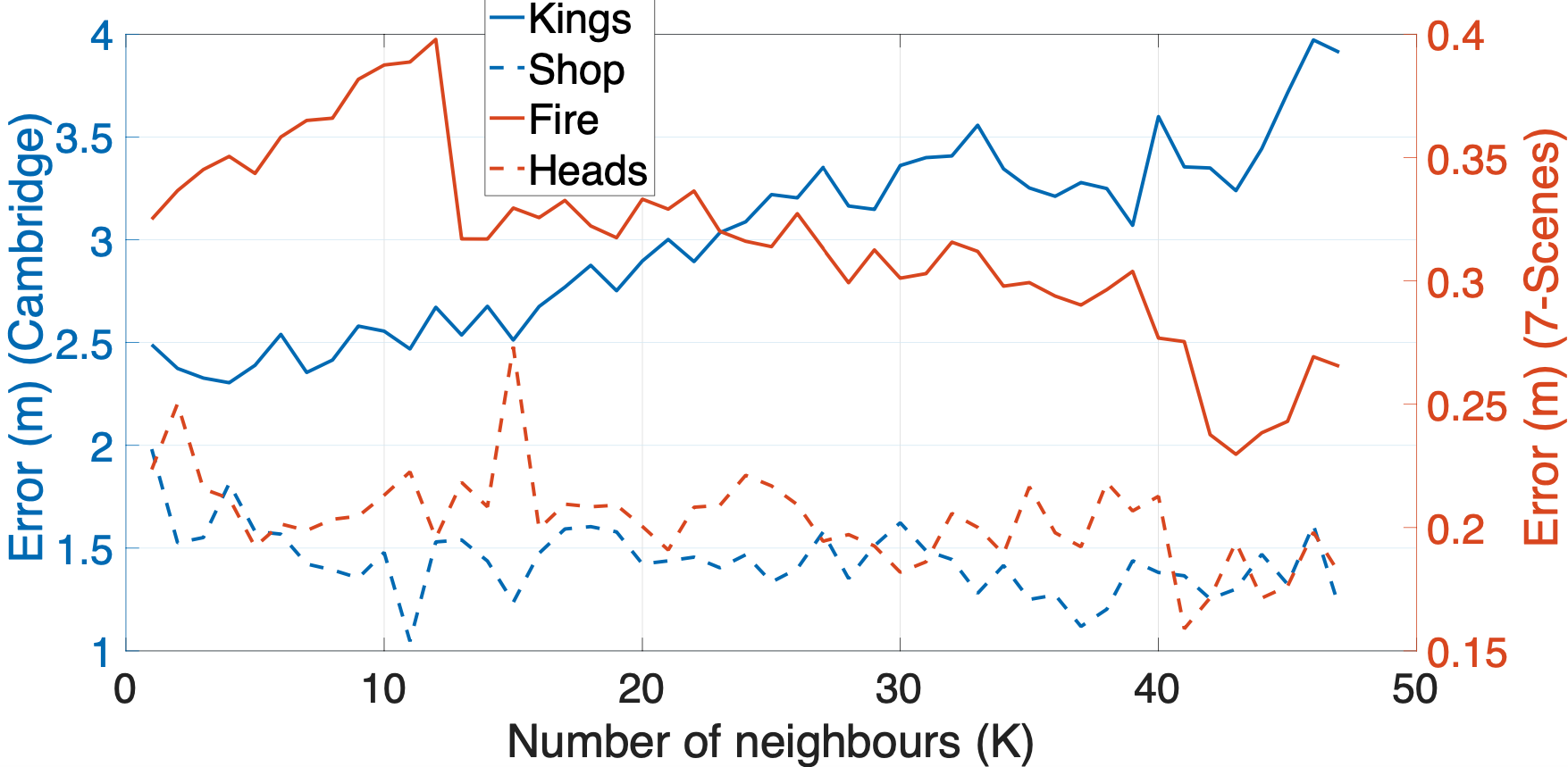}}

  \caption{The relation between the number of neighbors and the median position error.}
  \label{k} 
\end{figure*}

\subsection{Effect of the number of neighbors}
Both of the proposed GNN systems rely on aggregating the information between the central node and its neighbors where the number of neighbors is a hyperparameter we need to tune. Fig. \ref{k} shows the relation between the number of neighbors and median position error for the 7 scenes dataset (Heads and Fire scenes) on the right $y$ axis and the Cambridge dataset (Kings and Shop scenes) on the left $y$ axis for both the node-pose architecture (left) and graph-pose architecture (right). For clarity, the notion of neighbors differs for the two systems. The neighbors for the node-pose system are images with similar pretrained features. However, for pose-graph, the neighbors are similar elements from the pretrained features map. Firstly, for the node-pose system, we can see in Fig. \ref{k} that in the indoor Heads and Fire scenes, more than 35 neighbors are needed to reach the lowest positional error while less than 10 neighbors are used to achieve the best accuracy for the outdoor Kings and Shop scenes. One reason for such phenomenon is that the distance between consecutive images for the outdoor scenes is much higher than the indoor scenes. This means more neighbors for outdoor scenes will be far from each other which will not help localization. However, for the indoor scenes, the camera moves very slowly and adding more neighbors up to a certain number will be beneficial and help localization as they will be close to the central node. Moreover, Fig. \ref{neigbours_node} shows the same observation that the neighbors for the indoor scenes are closer than the outdoor scenes. In Fig. \ref{neigbours_node}, $2$ nodes from the training set and $2$ nodes from the test set for each scene along with their neighbors from the training set are visualized. In the case of the Heads scene, we can see that the positions of training nodes (in cyan and yellow) are very close which validates our claim that similar pretrained features of images leads to close positions.  However, for the test nodes, as we find their nearest neighbors from the training set, we can see that the neighbors are not as ideal as the training nodes but they lie on the same area of the testing node. For the Kings outdoor scene, although the connected positions are not very far from each other, they are not very close either contrary to the indoor scene case. This happens because the overlap between images is lower than the indoor scene and the distance between consecutive images is higher. 

Secondly, despite the different neighbors representation for the graph-pose system, the same behaviour is present as shown in Fig. \ref{k} where using more than 40 neighbors gives the best performance for indoor scenes and the lowest error for outdoor scenes is achieved using less than 10 neighbors. This behaviour is preserved since for indoor images, the field of view of images is small which can enable us to relate more pixels or features together than the outdoor images whose field of view is very wide and lower features from the images can be related as neighbors.

%\subsection{Effect of the intermediate feature maps}

\section{Performance analysis}
We validate the performance of the two proposed systems with the state of the art single image localization systems for all the scenes of the 7 scenes dataset and the Cambridge dataset interms of median position error (m) and median orientation error (deg) using both ImageNet and Places pretrained datasets in Table \ref{Median Error}. As seen, the node-pose system outperforms all the other methods for the 7 scenes dataset for both position and orientation reaching average error of $0.18 m/7.5 ^{\circ}$ using ImageNet pretrained features. Moreover, node-pose outperforms all the other methods using Places pretrained features for the orientation calculation of the outdoor scenes with an average error of $3.76^{\circ}$ and reaches comparable results to PN L-W And LSTM. PN for the position error with only few centimeters difference and outperforming all the other methods. We can notice the same observation mentioned in Subsection \ref{observations} that using Places dataset is better for outdoor scenes while ImageNet pretrained features perform better for the indoor environments. For graph-pose system, we observe that although node-pose system outperforms it, the graph-pose system still outperforms all the other methods for the 7 scenes dataset. However, for the Cambridge outdoor dataset, the graph-pose architecture does not perform as well as for the indoor scenes but produce comparable results to the other localization methods. 
\label{ex}
\section{Conclusion} 

%We propose a novel image based localization system using graph neural networks (GNN). We use pretrained ResNet50 %convolutional neural network (CNN) architecture to extract the important features for each image. Following, we use the %extracted features as input to GNN to find the pose of each image by either using the image features as a node in a %graph and formulate the pose estimation problem as node pose regression or modelling the image features themselves as a %graph and the problem becomes graph pose regression. We do an extensive comparison between the proposed two approaches %and the state of the art single image localization methods and we show that using GNN leads to enhancing the performance %for both indoor and outdoor environments. 

In this paper, a novel image based localization system using graph neural networks has been proposed. Image features from the pretrained ResNet50 CNN architecture are used as either node features in a graph or as a one graph.  Experiments for indoor and outdoor scenes have been done using multiple pretrained datasets and multiple types of GNNs. It has been found that representing images as nodes in a graph and learning the pose from neighbouring nodes leads to lower localization error for indoor and outdoor scenes than representing images as a graph. Moreover, using the central node features with a different set of weights than its neighbours can enhance the training and produces better results. Finally, for outdoor scenes, features from the Places dataset perform the best while features from ImageNet dataset produce the best results for indoor environments.

\section*{Acknowledgments}
This study was supported by the Nature Science and Engineering Research Council of Canada (NSERC) Collaborative Research and Development Grant (Grant No. CRDPJ-52098-2017), NSERC Discovery Grant (Grant No. RGPIN-2015-06515), and NVidia
Corporation TITAN-X GPU grant.

\bibliographystyle{IEEEtran}
\bibliography{refs.bib}{}
\end{document}